\documentclass[11pt]{article}

\usepackage[final]{acl}
\usepackage{times}
\usepackage{latexsym}

\usepackage[T1]{fontenc}

\usepackage[utf8]{inputenc}

\usepackage{microtype}

\usepackage{inconsolata}

\usepackage{graphicx}

\usepackage{amssymb}
\usepackage{adjustbox}     
\usepackage{hyperref}
\usepackage{microtype}
\usepackage{amsmath}
\usepackage{lipsum}
\usepackage{amsfonts}
\usepackage{subcaption}
\usepackage{comment}
\usepackage{url}
\usepackage{multirow}
\usepackage{booktabs}
\DeclareUnicodeCharacter{2212}{-}
\usepackage{makecell}
\usepackage{float}
\usepackage{array} 
\usepackage{xcolor}

\title{Reasoning-Based Refinement of Unsupervised Text Clusters with LLMs}

\author{Tunazzina Islam \\
  Department of Computer Science \\
  Purdue University \\
  West Lafayette, IN 47907 \\
  \texttt{islam32@purdue.edu} \\}

\begin{document}
\maketitle
\begin{abstract}
 Unsupervised methods are widely used to induce latent semantic structure from large text collections, yet their outputs often contain incoherent, redundant, or poorly grounded clusters that are difficult to validate without labeled data. We propose a \textbf{reasoning-based refinement framework} that leverages large language models (LLMs) not as embedding generators, but as semantic judges that validate and restructure the outputs of arbitrary unsupervised clustering algorithms. Our framework introduces three reasoning stages: (i) \textbf{coherence verification}, where LLMs assess whether cluster summaries are supported by their member texts; (ii) \textbf{redundancy adjudication}, where candidate clusters are merged or rejected based on semantic overlap; and (iii) \textbf{label grounding}, where clusters are assigned interpretable labels through a two-stage process that generates and consolidates semantically similar labels in a fully unsupervised manner. This design decouples representation learning from structural validation and mitigates the common failure modes of embedding-only approaches. We evaluate the framework in real-world social media corpora from two platforms with distinct interaction models, demonstrating consistent improvements in cluster coherence and human-aligned labeling quality over classical topic models and recent representation-based baselines. Human evaluation shows strong agreement with LLM-generated labels, despite the absence of gold-standard annotations. We further conduct robustness analysis under matched temporal and volume conditions to assess cross-platform stability. Beyond empirical gains, our results suggest that LLM-based reasoning can serve as a general mechanism for validating and refining unsupervised semantic structure, enabling more reliable and interpretable analysis of large text collections without supervision. 
\end{abstract}

\section{Introduction}

Large text collections are commonly analyzed using unsupervised methods to induce latent semantic structure, enabling downstream tasks such as summarization, monitoring, and social analysis without requiring labeled data \citep{angelov2020top2vec,hoyer2004non,blei2003latent,lee1999learning,deerwester1990indexing}. Such methods are particularly important in domains where annotation is costly or infeasible, including social media, where discourse is noisy, rapidly evolving, and highly heterogeneous \cite{zappavigna2012discourse}. Despite their widespread use, unsupervised clustering pipelines often produce outputs that are noisy, redundant, or weakly interpretable, making it difficult to assess whether the induced structure meaningfully reflects the underlying semantics of the data.
\begin{figure}[t]
\includegraphics[width=\columnwidth]{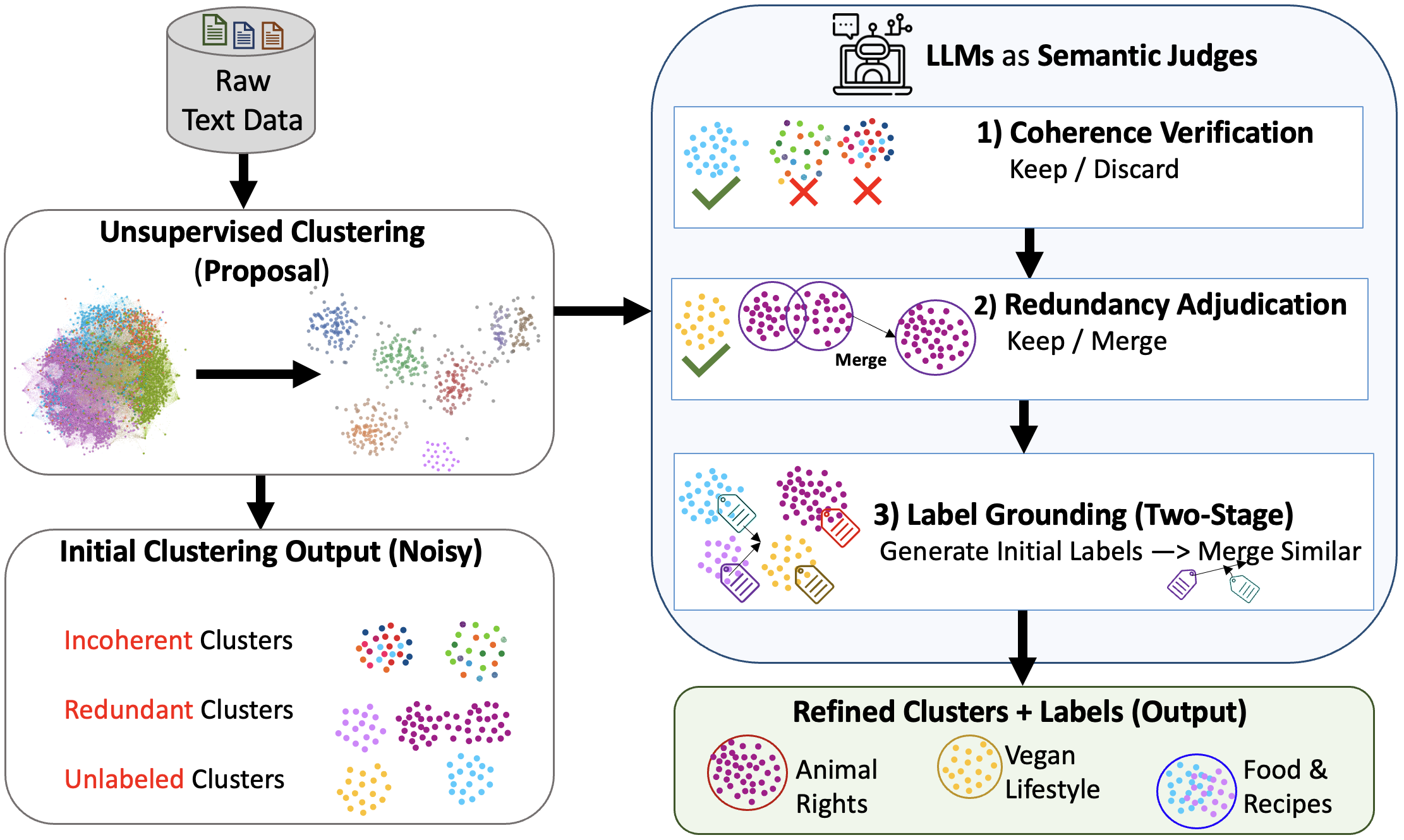}
\caption{{\small Overview of our framework. Unsupervised clustering generates initial cluster proposals that are often noisy. We treat these clusters as hypotheses and use LLMs as semantic judges to (1) verify coherence, (2) adjudicate redundancy, and (3) generate interpretable labels via a two-stage grounding process, producing refined, coherent, and distinct clusters.}} 
\vspace{-5pt}
\label{fig:frame}
\vspace{-5pt}
\end{figure}

A central challenge in unsupervised semantic structure induction is that the number of latent themes is typically \textbf{unknown a priori}. To address this, prior work has frequently adopted non-parametric formulations that allow the structure to grow with the data \cite{srijith2017sub}, for example, through hierarchical Bayesian models such as the Hierarchical Dirichlet Process (HDP) \cite{teh2004sharing}. While such approaches mitigate the need to pre-specify the number of clusters, they do not resolve a more fundamental issue: whether the induced clusters are semantically coherent, non-redundant, and interpretable to humans, especially in short-text and high-noise settings such as social media.

More recent approaches improve unsupervised text clustering by learning stronger representations, typically using contextual sentence embeddings combined with geometric clustering criteria \citep{grootendorst2022bertopic,reimers2019sentence}. These methods assess cluster quality through distance-based properties of the embedding space, such as separation or density, treating them as indicators of semantic coherence. However, prior work has shown that embedding geometry does not always align with human notions of meaning \citep{ethayarajh-2019-contextual, mimno2011optimizing}. In practice, clusters can appear well separated numerically while remaining semantically incoherent, and multiple clusters may encode overlapping themes with only superficial lexical differences. As a result, representation-centric pipelines lack explicit mechanisms for verifying whether induced clusters are meaningful or interpretable from a semantic perspective.


In this paper, we take a different approach. Rather than proposing a new clustering algorithm or representation, we study how large language models (LLMs) \cite{brown2020language} can be used as semantic reasoners to validate and restructure the outputs of arbitrary unsupervised clustering methods. Our key insight is that LLMs possess strong natural-language reasoning capabilities \citep{yao2022react,wei2022chain, kojima2022large} that can be leveraged to assess whether a proposed cluster is internally coherent, whether two clusters are meaningfully distinct, and whether an induced theme is well-grounded in the underlying texts. This enables a shift from purely statistical or geometric criteria toward explicit semantic validation.

We introduce a \textbf{reasoning-based cluster refinement} framework (Fig. \ref{fig:frame}) that treats clustering as a \textbf{proposal step} and uses LLM reasoning to adjudicate structure. The framework consists of \textbf{three} stages:
\newline
 (1) coherence verification, where LLMs assess whether a cluster summary is supported by its member texts;
 \newline
 (2) redundancy adjudication, where candidate clusters are merged or rejected based on semantic overlap rather than embedding similarity alone; and
 \newline
 (3) label grounding, where clusters are assigned interpretable labels through a \textbf{two-stage} process that generates candidate labels and consolidates semantically similar ones.
 
 Importantly, LLMs are used not as embedding generators, but as \textbf{semantic judges} that accept, reject, or revise structural hypotheses produced by unsupervised methods.
 Unlike topic modeling approaches that induce semantic structure from scratch, our framework treats clustering output as a hypothesis and uses LLM reasoning only to validate, prune, and ground that structure.
 This design decouples representation learning from structural validation, mitigating common failure modes of embedding-only pipelines. Our framework is agnostic to the choice of clustering algorithm and can be applied as a post-hoc refinement layer to existing unsupervised systems.

We evaluate the proposed framework on large-scale social media corpora drawn from two platforms with distinct interaction models: X (formerly Twitter) and Bluesky. While the empirical study focuses on \textit{vegan} discourse—a socially impactful and contested topic—the domain serves primarily as a \textbf{testbed} for evaluating the framework under realistic noise, redundancy, and platform variation. 
To assess reliability in the absence of gold annotations, we conduct a human evaluation with expert annotators, achieving high inter-annotator agreement and demonstrating strong alignment between LLM-generated labels and human interpretations. We further perform robustness analysis under matched temporal and volume conditions to examine the stability of the induced structure across platforms. 
Our contributions are threefold:
\begin{enumerate}\setlength{\itemsep}{0pt}\setlength{\parskip}{0pt}
    \item We propose a reasoning-based framework for validating and refining the unsupervised semantic structure using LLMs as \textbf{semantic judges}.
    \item We provide a systematic evaluation, including human validation, comparing reasoning-based refinement with embedding-only approaches.
    \item We release cross-platform datasets and evaluation resources to support future research on interpretable and reliable unsupervised text analysis\footnote{Our datasets and code are available here: \href{https://github.com/tunazislam/reasoning-based-refinement-llms-vegan}{https://github.com/tunazislam/reasoning-based-refinement-llms-vegan}}.
\end{enumerate}

\section{Related Work}
A large body of previous work studies unsupervised methods for inducing latent semantic structure from text \cite{boyd2014care,blei2003latent}. Classical probabilistic approaches such as Latent Dirichlet Allocation (LDA) \cite{blei2003latent} and its non-parametric extensions, including HDP \cite{teh2004sharing}, address the problem of unknown cluster cardinality by allowing the number of latent components to grow with the data. While these methods provide principled statistical formulations, they often struggle with semantic coherence and interpretability, particularly in short and noisy texts such as social media posts \citep{mimno2011optimizing,hong2010empirical,chang2009reading}.

More recent approaches leverage representation learning to improve unsupervised structure induction. Contextual sentence embeddings \citep{reimers2019sentence} combined with density-based or hierarchical clustering have become common, exemplified by methods such as BERTopic \citep{grootendorst2022bertopic}. These approaches improve cluster separation in the embedding space, but fundamentally rely on geometric criteria as proxies for semantic validity. As a result, clusters may appear statistically well-formed while remaining semantically incoherent, redundant, or weakly grounded in natural language \citep{ethayarajh-2019-contextual}. Our work departs from representation-centric approaches by focusing on \emph{validating} induced structure rather than improving representations.

Unsupervised structure induction has been widely applied to social media data to study public opinion, discourse dynamics, and emerging narratives \citep{momeni2018modeling,zhao2011comparing}. However, social media text presents persistent challenges for unsupervised methods, including short length, high lexical variation, sarcasm, and rapid topical drift. These properties exacerbate the gap between statistical coherence and human interpretability, limiting the reliability of downstream analyses based on automatically induced themes.
Rather than proposing domain-specific heuristics or new topic models, our approach treats clustering outputs as \emph{hypotheses} that require semantic validation. 

LLMs have recently been used to support data annotation, weak supervision, and qualitative analysis across a range of NLP tasks \cite{wang2023large,ding2023gpt,wang2021want}. Prior work demonstrates that LLMs can generate labels, summaries, and explanations that align closely with human judgments, enabling scalable annotation in low-resource settings \citep{islam2025can,gilardi2023chatgpt,huang2023chatgpt}. These capabilities have motivated the use of LLMs for theme labeling and content summarization in social science and computational social science research.
Most existing approaches, however, employ LLMs as generative annotators or topic
induction tools, applying them directly to texts or clusters to produce structure \cite{brady2025latent,pham2023topicgpt}. \citet{islam-goldwasser-2025-post} used LLM-generated explanations to extract recurring themes and aspects. Unlike recent approaches that use LLMs to directly generate topics or cluster structures, our framework instead uses LLMs as \textit{semantic judges} to validate, refine, and ground the structure produced by unsupervised methods.



A growing line of work explores LLMs-in-the-loop for improving unsupervised or weakly supervised systems \citep{islam2025uncovering,islam2025discovering,dai2023llm}. \citet{islam2025discovering} guided their framework using a seed set of initial themes. Building on this, \citet{islam2025uncovering} assumed a predefined set of themes and focused on uncovering underlying arguments. \citet{lam2024concept} developed a concept induction algorithm using LLM with human-guided abstraction called LLooM, which has a \textit{seed} operator for accepting user-provided seed term/set. In contrast, our method operates without any initial seed set.

\section{Problem Formulation}

Let $\mathcal{D} = \{x_1, \ldots, x_N\}$ denote a corpus of unstructured text documents, such as social media posts. The goal of unsupervised semantic structure induction is to organize $\mathcal{D}$ into a set of latent semantic groupings that support downstream analysis and interpretation, without access to labeled data.

In practice, existing unsupervised pipelines typically proceed by applying a clustering algorithm to learned text representations, producing a set of clusters $\mathcal{C} = \{C_1, \ldots, C_K\}$, where $K$ is not known a priori. We refer to each $C_k$ as a \emph{cluster hypothesis}: a candidate grouping proposed by an unsupervised method that may or may not correspond to a coherent semantic theme.

Our focus is not on generating cluster hypotheses, but on addressing a complementary and underexplored problem: \emph{how to validate and refine cluster hypotheses in the absence of labeled data}. Specifically, given an initial set of clusters $\mathcal{C}$, we aim to produce a refined semantic structure $\mathcal{C}^\ast$ that satisfies three design goals:
(i) \textbf{semantic coherence}, where documents within a cluster support a common theme;
(ii) \textbf{non-redundancy}, where distinct clusters correspond to meaningfully different themes; and
(iii) \textbf{interpretability}, where each cluster can be grounded in a concise, human-readable description, potentially through consolidation of semantically overlapping label candidates.


\subsection{Design Principles}
Our framework is guided by three design principles.

\noindent\textbf{Clustering as proposal.} We treat unsupervised clustering as a proposal mechanism that generates candidate structure, rather than as a final decision. The choice of clustering algorithm is therefore interchangeable and not central to the framework.

\noindent \textbf{LLMs as semantic judges.} Large language models are used not to generate embeddings or clusters, but to reason over natural-language summaries and textual evidence. Given a cluster hypothesis, an LLM evaluates whether the hypothesis is semantically supported by its member documents.

\noindent \textbf{Explicit reasoning checkpoints.} 
We decompose validation into explicit reasoning stages that target specific failure modes of unsupervised clustering, such as incoherence.


\subsection{Framework Overview}

Fig.~\ref{fig:frame} illustrates the overall pipeline. The framework takes as input an initial set of cluster hypotheses $\mathcal{C}$ produced by any unsupervised clustering method, along with the underlying documents.

\noindent\textbf{Stage 1: Coherence Verification.}
For each cluster hypothesis $C_k$, we first construct a concise natural-language summary that captures its latent theme. An LLM then evaluates whether this summary is supported by the documents in $C_k$, reasoning over representative examples. Clusters deemed semantically incoherent are discarded from the final cluster set.

\noindent\textbf{Stage 2: Redundancy Adjudication.}
Even when clusters are individually coherent, multiple clusters may encode overlapping or redundant themes. To address this, we compare summaries of surviving clusters pairwise, and redundant clusters are merged, while genuinely distinct clusters are retained.

\noindent\textbf{Stage 3: Label Grounding.}
For each refined cluster in $\mathcal{C}^\ast$, 
the LLM assigns interpretable labels through a \textbf{two-stage} grounding process, where initial candidate labels are generated, and semantically similar labels are consolidated.


The output of the framework is a refined set of clusters $\mathcal{C}^\ast$ with grounded labels, which can be evaluated using both automatic metrics and human judgment, as described in subsequent sections.

\section{Experiments}
In this section, we detail the dataset, experimental setup to implement our framework, results, and error analysis.
\vspace{-5 pt}
\subsection{Dataset}
Prior studies of online discourse on lifestyle choices have largely focused on single, centralized platforms such as X \cite{islam2019yoga}. In contrast, we include vegan discourse from both \textbf{X} and \textbf{Bluesky}, an emerging decentralized social network, to evaluate our framework across platforms with different interaction dynamics.

To extract tweets and posts regarding vegan lifestyle choices, we filter texts containing specific keywords, such as \textit{vegan}, \textit{veganism}, \textit{plantbased}, \textit{meatfree}. The full list of keywords is shown in Table \ref{tab:keywords} in App. \ref{app:dc}. 
For X, we collect tweets using the Tweepy\footnote{\url{https://www.tweepy.org/}} library via the Twitter streaming API sub-sequentially from October 2019 to February 2020. We have extracted $330464$ tweets from $204670$ different users. Later, we notice that there are $63751$ suspended users. However, in this work, we did not use the whole dataset. To avoid dominance by prolific or automated accounts, we sample at the user level, limiting the number of posts per user. Finally, we have $20000$ tweets from $275$ different users. This design prioritizes thematic diversity over volume and follows common practice in social media discourse analysis.

For Bluesky, we use a lightweight pipeline to collect and store the posts from the Bluesky firehose\footnote{\url{https://docs.bsky.app/docs/advanced-guides/firehose}} in real time. It consists of a \textit{data collector} that connects to the firehose and collects the new posts. We \textbf{do not} download \textit{comments} of corresponding posts. We have collected $13032$ English posts from June 2025, of which $1752$ are unique.

\paragraph{Disclaimer:} The datasets analyzed in this study consist of publicly available posts from X and Bluesky. These data may contain offensive content or toxic language. The content is used solely for academic analysis and not for the dissemination of harmful material.
\subsection{Framework Implementation}
\subsubsection{Clustering Texts}
For the clustering process, we utilized a multi-step approach combining dimensionality reduction, clustering, and hyperparameter search optimization. We begin by clustering the texts using HDBSCAN \cite{mcinnes2017hdbscan}, a nonparametric clustering algorithm. 
First, the text data is pre-processed and transformed into numerical form using TF-IDF vectorization. While dense embedding models such as GTR-T5 \cite{ni2022large}, E5 \cite{wang2022text} could serve as alternatives, we adopt TF–IDF to preserve interpretability and lexical transparency in early clustering stages, ensuring that the LLM-based refinement focuses on semantic consolidation rather than embedding quality. We then normalize the sparse matrix representation using MaxAbsScaler to ensure balanced scaling across features. To reduce the dimensionality, Truncated Singular Value Decomposition (SVD) is applied, capturing the most important information from the data in a lower-dimensional space. The UMAP \cite{mcinnes2018umap} is employed to further reduce dimensions and create embeddings that preserve the local structure of the data. The HDBSCAN is optimized by searching across different parameter configurations (e.g., min\_cluster\_size, min\_samples, cluster\_selection\_method, metric) using the DBCV \citep{moulavi2014density} score as an evaluation metric. 


\subsubsection{Refining the Clusters}
\begin{figure}[h]
\includegraphics[width=\columnwidth]{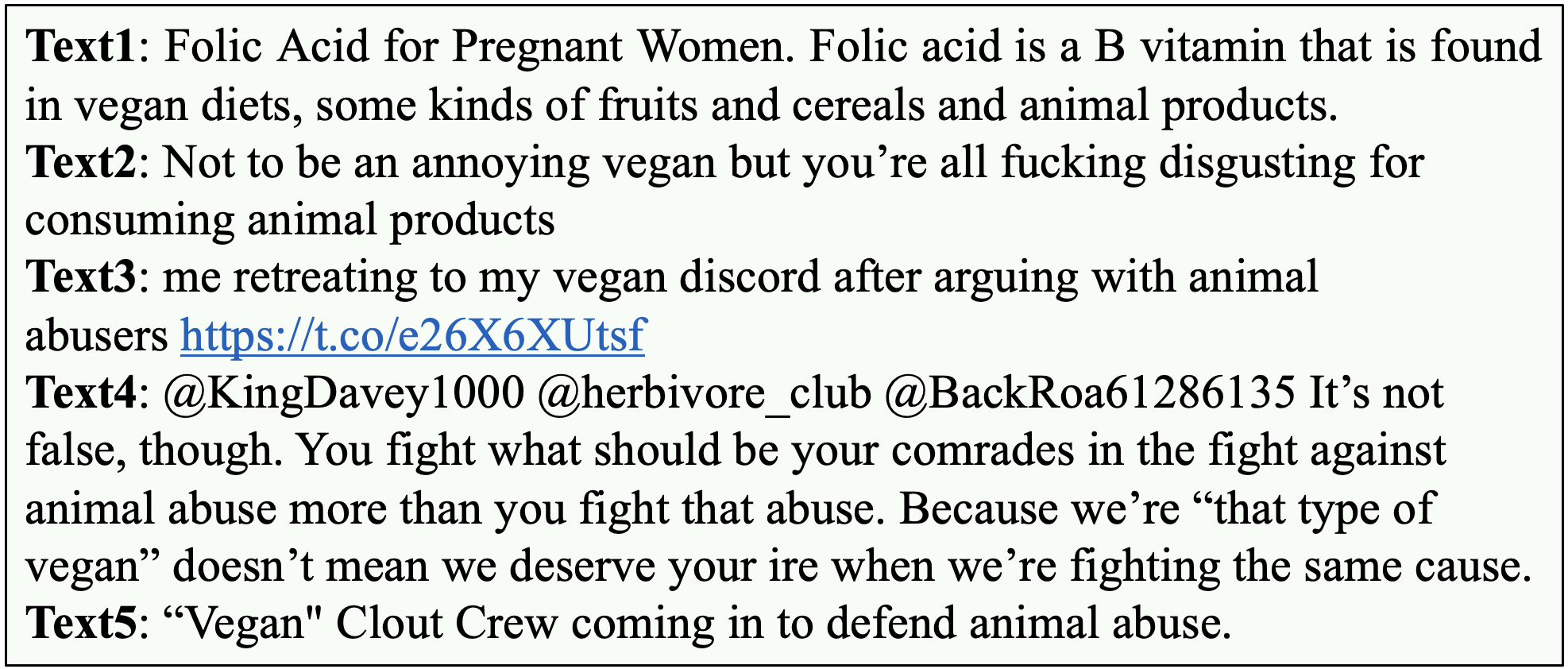}
\caption{{\small Example of the incoherent cluster from X.}}
\vspace{-5pt}
\label{fig:incoh}
\vspace{-5pt}
\end{figure}

\begin{figure}[h]
\centering
\begin{subfigure}{.95\columnwidth}
  \centering
  \includegraphics[width=1\textwidth]{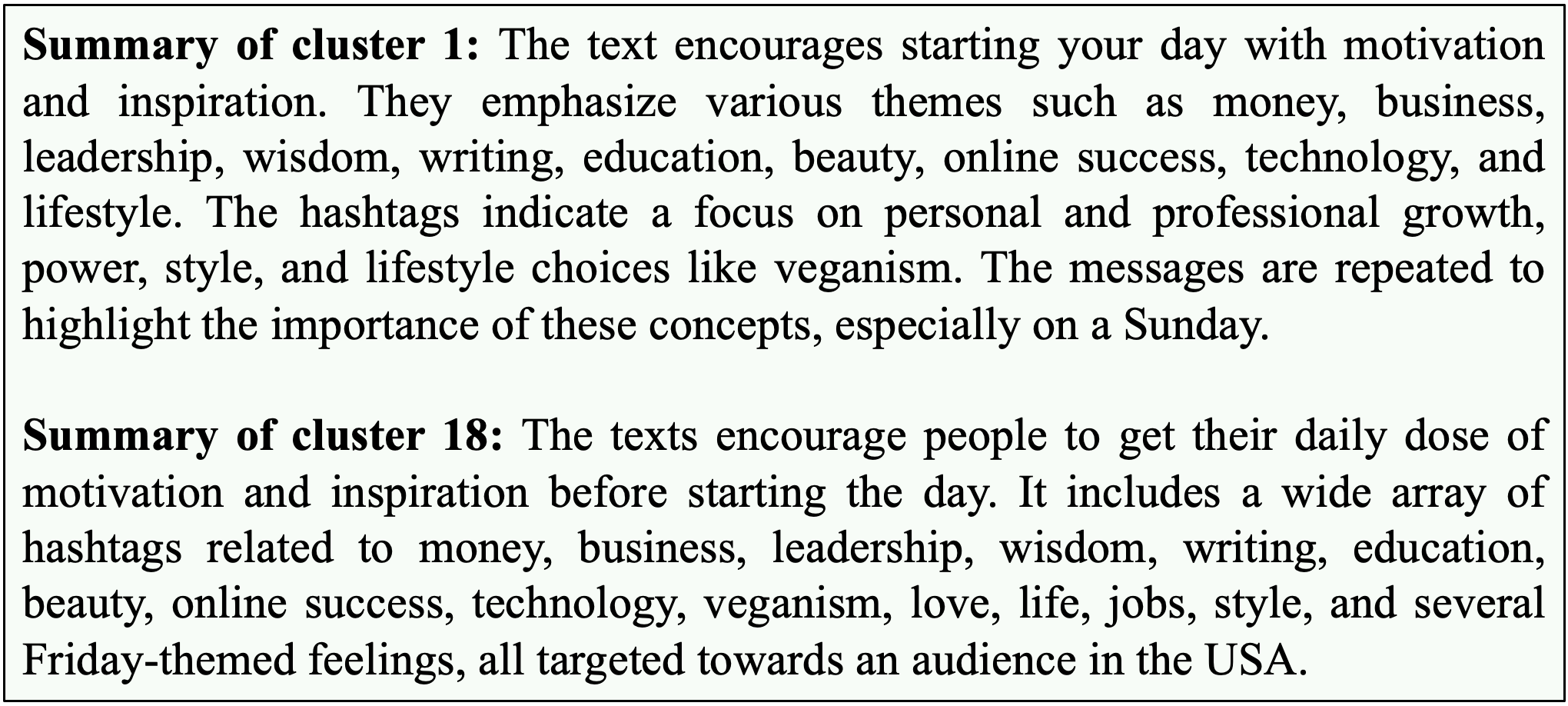}
  \caption{{\small Merged clusters (X).}}
  \label{fig:merge_x}
\end{subfigure}
\begin{subfigure}{.95\columnwidth}
  \centering
  \includegraphics[width=1\textwidth]{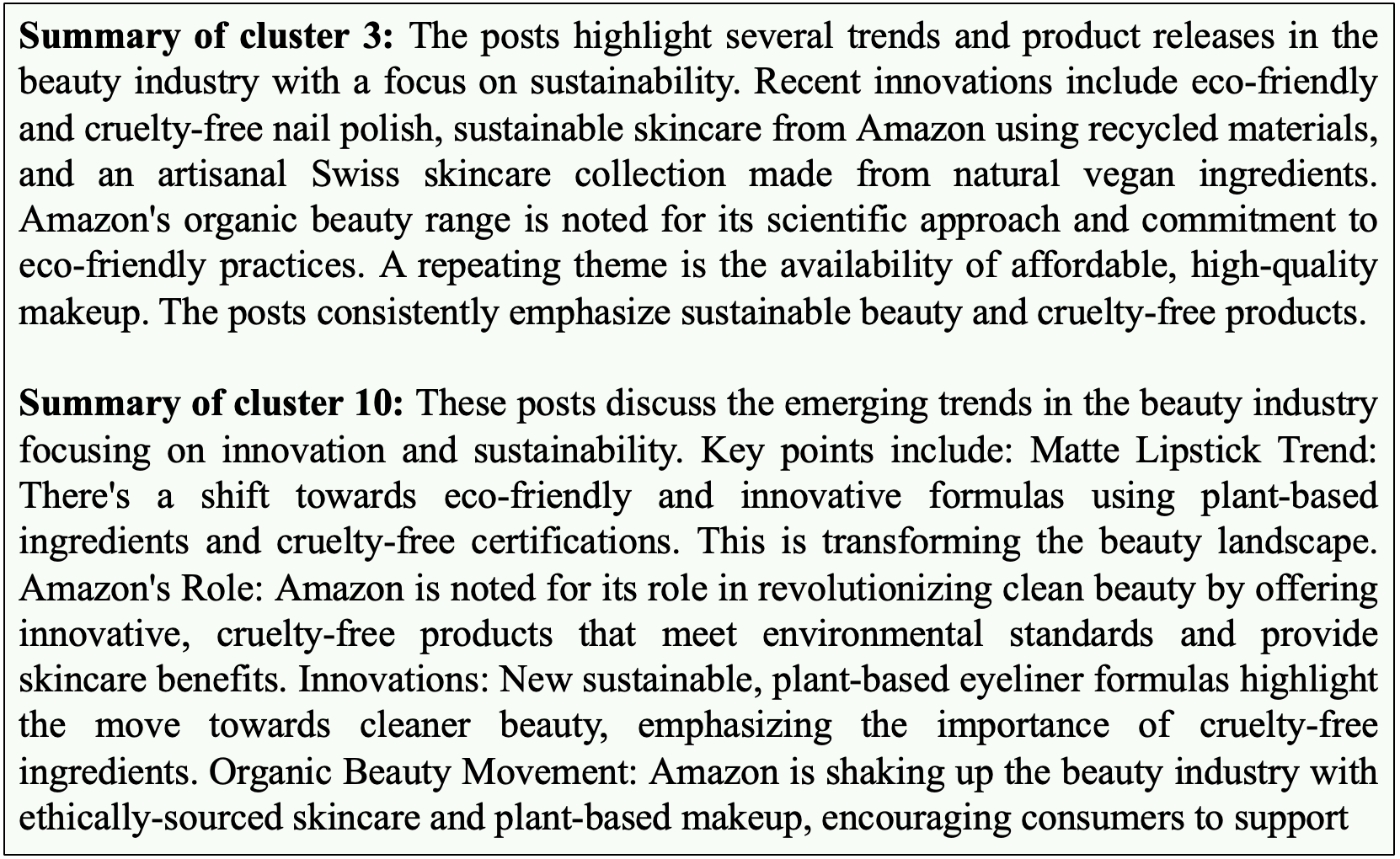}
  \caption{{\small Merged clusters (Bluesky).}}
  \label{fig:merge_blusky}
\end{subfigure}
\vspace{-5pt}
\caption{{\small Example of merged cluster summaries.}}
\vspace{-5pt}
\label{fig:merge}
\vspace{-7pt}
\end{figure}

To enhance the semantic coherence of the clusters obtained from HDBSCAN, we perform a multi-step refinement process involving LLMs and Sentence-BERT (SBERT) \citep{reimers2019sentence}.
\vspace{-5pt}
\paragraph{Generating Cluster Summaries.}
For each cluster, we utilize LLM with a zero-shot prompting manner to generate a concise summary from top-k (k = $5$) texts assigned to each cluster. 
To construct representative summaries, we select the \textbf{top-5 documents closest to the cluster centroid in embedding space}. This choice reflects a trade-off between representativeness and prompt stability. In pilot experiments, we observe that cluster summaries stabilize beyond $k=5$, with minimal qualitative changes when additional documents are included. Using a small representative set also helps avoid prompt dilution when clusters contain many heterogeneous texts, ensuring that the most central examples guide the summary generation. Prior work on LLM-assisted discourse analysis \cite{islam2025uncovering,islam2025discovering} adopts a similar representative-sampling strategy. We also verified robustness across $k \in \{3,5,7\}$, observing qualitatively consistent summaries and coherence judgments. Finally, limiting the prompt size reduces latency and API cost while maintaining stable cluster descriptions.
\vspace{-5pt}
\paragraph{Verifying Cluster Coherence.}
For coherence verification, the LLM evaluates the semantic alignment between the generated cluster summary and the representative texts used to construct it. The evaluation operates on these representative samples rather than the full set of cluster documents. A cluster is flagged as incoherent when the summary fails to capture a consistent theme across multiple representative texts. In this context, \textit{low semantic alignment} refers to cases where the LLM determines that the summary is not sufficiently supported by the representative documents, indicating that the cluster may contain heterogeneous or weakly related content.
Fig. \ref{fig:incoh} shows an example of an incoherent cluster from our dataset.
\vspace{-7pt}
\paragraph{Identifying and Merging Redundant Clusters.}
We identify clusters that are semantically similar and merge them to reduce redundancy. SBERT is used to generate an embedding for each cluster summary. Cosine similarity between cluster embeddings is calculated to identify clusters with high semantic overlap. Clusters exceeding a similarity threshold are merged. We select the appropriate threshold using grid search. Fig. \ref{fig:merge} in 
shows examples of merged cluster summaries from X (Fig. \ref{fig:merge_x}) and Bluesky (Fig. \ref{fig:merge_blusky}) datasets.

\subsubsection{Generating Cluster Label}
Each cluster initially produces a single candidate label derived from its summary. After this step, we compute pairwise SBERT similarity between generated labels. Labels with similarity above a threshold of $0.85$ are grouped to identify semantically redundant themes. For each group of similar labels, the LLM generates a consolidated label representing the merged semantic category. This process allows multiple preliminary labels to be unified into a single final label (Table \ref{tab:label} in App. \ref{app:gen_label}), forming a \textbf{two-stage labeling} procedure consisting of initial label generation followed by semantic consolidation.
\subsubsection{Assigning Label to Individual Text}
After the final set of consolidated labels is produced, individual documents are reassigned using LLM to ensure alignment between the refined semantic taxonomy and document-level content.
This way, we assign the generated label to each text. The LLMs assign the most appropriate label to each text based on its content. Prompts are designed to guide the LLMs in selecting the best-fitting label from the set of generated labels.


\subsection{Experimental Details}
For the HDBSCAN clustering model, we use a data-driven approach to estimate the best number of topics by maximizing the DBCV score. We retain the default
settings for $cluster\_selection\_method$, and $metric$
parameters, while we grid search the $min\_samples$ and $min\_cluster\_size$ 
to get more sensible topics (Detail in App. \ref{app:hdbscan}).


We grid search the merge similarity threshold $\tau \in \{0.75, 0.80, 0.85, 0.90\}$ using standard metrics: Silhouette Score ($S$, higher is better) \cite{rousseeuw1987silhouettes}, Davies–Bouldin Index ($DB_i$, lower is better) \cite{davies2009cluster}, and cluster count ($C$). We select $\tau = 0.85$ as the best trade-off—strong scores with controlled cluster count (see App. \ref{app:threshold_selection} and Table \ref{tab:thrs_grid} for grid search results).

For the LLM part of implementation, we use GPT-4o\footnote{\url{https://openai.com/index/hello-gpt-4o/}} \cite{openai2024gpt4o} with the default parameters. 
%
Finally, we have $14$ and $22$ generated labels for X and Bluesky, respectively, after following the steps mentioned in our framework. The generated labels are shown in Table \ref{tab:label} in App. \ref{app:gen_label}. Label distributions are detailed in App. \ref{app:thm_dis}. Prompt details are provided in App. \ref{app:prompt} in Fig. \ref{prompt_template} and Fig. \ref{prompt_ex}. Cost is provided in App. \ref{app:cost}.

\subsection{Baselines}
To evaluate whether applying the LLM-based refinement process increases the coherence of clusters generated by HDBSCAN, we measure cluster quality, semantic coherence, and conduct statistical significance testing. 
We have the following two baselines:
\newline
\textbf{HDBSCAN without Refinement}: Evaluating the original clusters from HDBSCAN.
\newline
\textbf{SBERT-Based Refinement}: Using SBERT for cluster refinement without LLM assistance. 

\subsubsection{Cluster Quality:}
Cluster quality evaluation is essential to verify that a clustering method produces meaningful groupings. To this end, we report both Silhouette Score and Davies–Bouldin Index, two widely used and complementary metrics.
As shown in Table~\ref{tab:cluster_coherency}, LLM-based refinement consistently
improves semantic coherence across both platforms, as measured by silhouette
scores. In contrast, SBERT-based refinement achieves stronger separation on X
according to $DB_i$, while differences are not statistically significant on Bluesky.
Overall, LLM-based refinement improves semantic coherence and human-aligned labeling
quality while achieving competitive separation compared to SBERT-based refinement. 

\begin{table}[h]
\centering
\small
\begin{tabular}{p{.82cm} | p{1.1 cm} | p{1.3 cm}| p{1.3 cm}| p{1 cm}}
\hline
Dataset & Metric & HDBSCAN & SBERT-rf & LLM-rf \\
\hline
\multirow{3}{*}{X} 
 & $C$ & 359 & 250 & 232 \\
 & $S$ ($\uparrow$) & 0.122 & 0.156 & \textbf{0.674} \\
 & $DB_i$ ($\downarrow$) & 2.322 & \textbf{0.569} & 0.635 \\
\hline
\multirow{3}{*}{Bluesky} 
 & $C$ & 37 & 34 & 36 \\
 & $S$ ($\uparrow$) & -0.017 & 0.052 & \textbf{0.979} \\
 & $DB_i$ ($\downarrow$) & 2.739 & 0.282 & \textbf{0.227} \\
\hline
\end{tabular}
\caption{{\small Cluster quality comparison. rf: refinement.}}
\vspace{-5pt}
\label{tab:cluster_coherency}
\vspace{-5pt}
\end{table}
\begin{figure}[h]
\centering
\begin{subfigure}{.5\columnwidth}
  \centering
  \includegraphics[width=\textwidth]{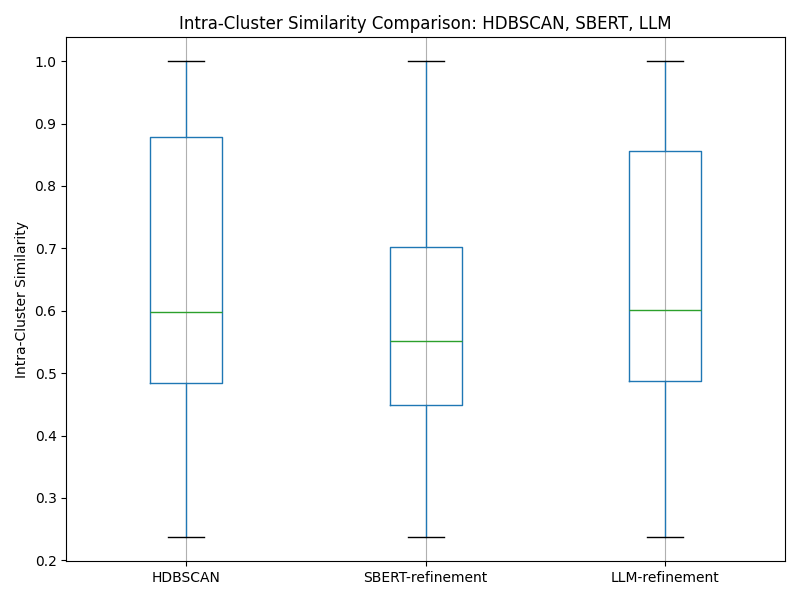}
  \caption{Box plot (X data).}
  \label{fig:x_intra_box}
\end{subfigure}%
\begin{subfigure}{.5\columnwidth}
  \centering
  \includegraphics[width=1\textwidth]{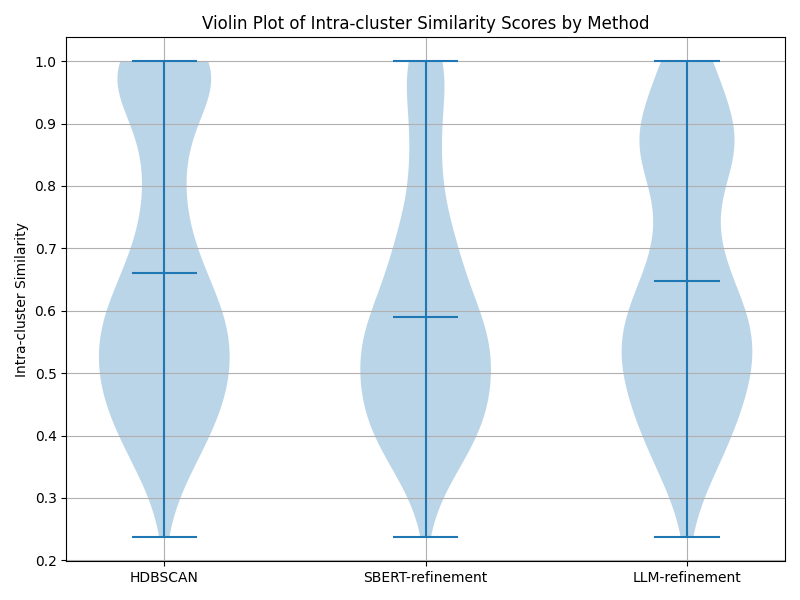}
  \caption{Violin plot (X data).}
  \label{fig:x_intra_violin}
\end{subfigure}
\begin{subfigure}{.5\columnwidth}
  \centering
  \includegraphics[width=1\textwidth]{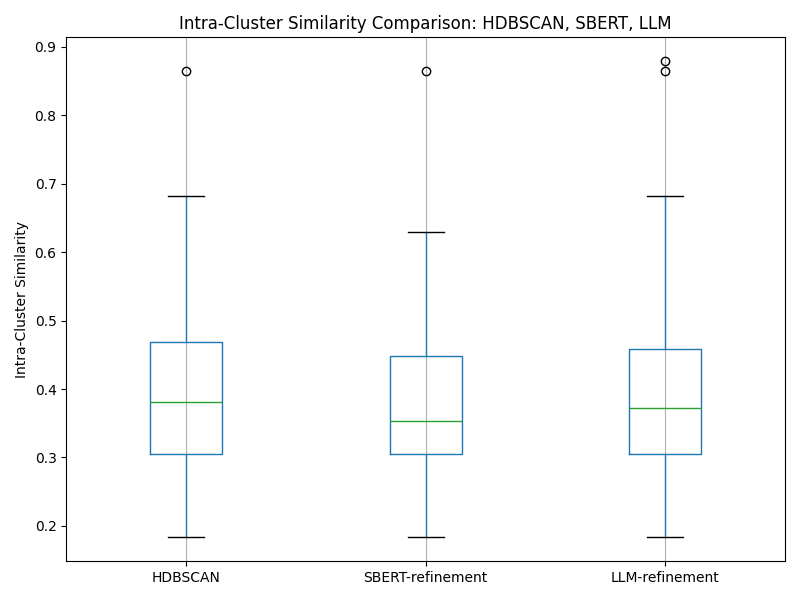}
  \caption{Box plot (Bluesky data).}
  \label{fig:blusky_intra_box}
\end{subfigure}%
\begin{subfigure}{.5\columnwidth}
  \centering
  \includegraphics[width=1\textwidth]{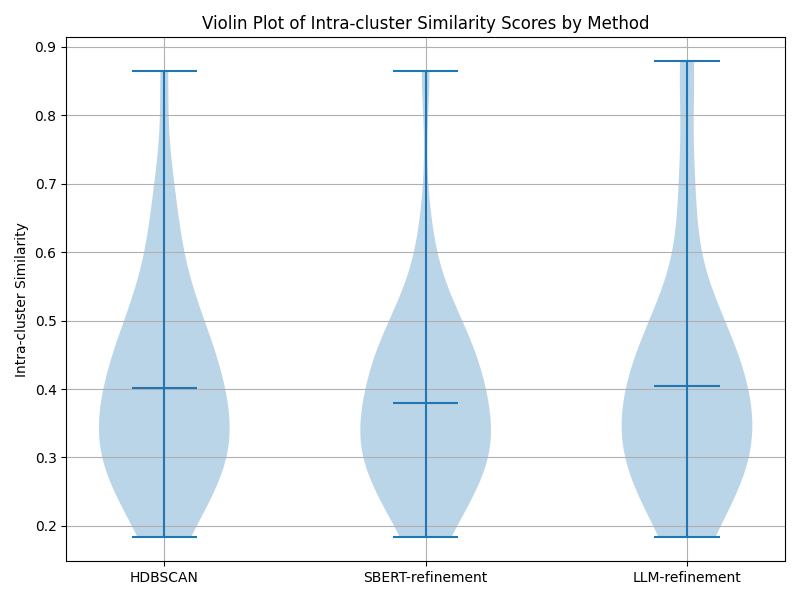}
  \caption{Violin plot (Bluesky data).}
  \label{fig:blusky_intra_violin}
\end{subfigure}
\vspace{-5pt}
\caption{{\small Comparing intra-cluster similarity across HDBSCAN, SBERT-refinement, and LLM-refinement.}}
\vspace{-5pt}
\label{fig:intra}
\vspace{-5pt}
\end{figure}
\subsubsection{Semantic Coherence:}
We measure intra‑cluster coherence as the mean cosine similarity of all text pairs in a cluster ($\tau\geq 0.85$ on sentence embeddings). Fig. \ref{fig:intra} plots the score distributions for X and Bluesky. 
On X data (Figs. \ref{fig:x_intra_box}, \ref{fig:x_intra_violin}), HDBSCAN and LLM‑refinement yield consistently higher similarity than SBERT‑refinement. Their medians are comparable ($\sim 0.60$), but LLM‑refinement slightly edges ahead. Both methods produce many clusters with very high cohesion ($0.9$–$1.0$), while SBERT‑refinement rarely does. 
On Bluesky (Figs. \ref{fig:blusky_intra_box}, \ref{fig:blusky_intra_violin}), HDBSCAN and LLM‑refinement again align, showing higher medians ($\sim 0.38$–$0.40$) and broader upper tails than SBERT‑refinement ($\sim 0.35$). All methods cluster most data in the moderate range ($0.3$–$0.5$), with occasional highly cohesive outliers; HDBSCAN and LLM‑refinement reach $\sim 0.87$–$0.89$, SBERT‑refinement lower.
\begin{table}[t]
\centering
\small
\begin{tabular}{p{2cm} | p{1.5 cm} | p{2.5 cm}}
\hline
Model & X Acc. (\%) & Bluesky Acc.(\%) \\ \hline
LDA & 30.4 & 36.2 \\ \hline
BERTopic & 38.7 & 42.4 \\ \hline
SBERT & 56.2 & 53.6 \\ \hline
TopicGPT & 72.8 & 68.4 \\ \hline
Llama 3.2 & 66.6 & 60.0 \\ \hline
Mistral Large 2 & 71.6 & 71.8 \\ \hline
\textbf{GPT-4o} & \textbf{78.4} & \textbf{89.8} \\ \hline
\end{tabular}
\caption{{\small Assignment comparison w.r.t. human judgment.}}
\vspace{-5pt}
\label{tab:eval}
\vspace{-5pt}
\end{table}
\subsubsection{Statistical Significance Test:}
To statistically test whether the coherence scores differ significantly between the initial and refined clustering, we compute the difference among the three methods: HDBSCAN w/o refinement, HDBSCAN with SBERT-refinement, and HDBSCAN with LLM-refinement. 
Table \ref{tab:stat} in App. \ref{app:stat} summarizes the statistical significance analysis of the clustering results for both the X and Bluesky vegan datasets. We first apply a non‑parametric Kruskal–Wallis \cite{kruskal1952use} test to determine whether there are overall differences in the clustering quality metrics.
For the X dataset, the Kruskal–Wallis test indicates a significant difference among the methods ($H=16.187$, $p<0.001$), prompting post‑hoc pairwise comparisons (independent samples) using the Mann–Whitney U \cite{mann1947test} test. These comparisons reveal that HDBSCAN with SBERT-refinement significantly outperformed both HDBSCAN ($p<0.001$) and LLM-refinement ($p<0.01$), while the difference between HDBSCAN and LLM-refinement is not statistically significant ($p=0.4808$).
For Bluesky, the Kruskal–Wallis test does not reveal significant differences across methods, and none of the Mann–Whitney U pairwise comparisons reached significance. 
\subsection{Evaluation}
We perform a human evaluation by asking annotators to assess whether the LLM-assigned theme adequately matches the content of \textbf{randomly} (random seed=$42$) selected $500$ tweets from X and $500$ posts from Bluesky datasets. Two annotators (experts in NLP and CSS) provide the judgment, and the inter-annotator agreement is $0.82$ (almost perfect agreement) using Cohen’s Kappa coefficient \cite{cohen1960coefficient}. We use multiple baselines, including SBERT-based assignment, topic modeling: LDA (with $10$ topics for X, $5$ topics for Bluesky), and BERTopic for comparison. For assignment, we compare three LLMs: GPT-4o, Mistral Large 2 (mistral-large-2407\footnote{\url{https://mistral.ai/news/mistral-large-2407/}}
) \citep{jiang2023mistral}, Llama 3.2 (llama-3.2-90b-text-preview\footnote{\url{https://www.llama.com/}}) \citep{touvron2023llama}. Table \ref{tab:eval} shows the human evaluation results on three LLMs for assignment as well as SBERT assignment and LDA, BERTopic baselines. LLM‑based labeling achieved high alignment with human judgments ($\sim 90\%$ on Bluesky), outperforming traditional and hierarchical topic modeling as well as SBERT baselines. This demonstrates that LLMs can serve as effective unsupervised annotators, enabling scalable theme assignment without costly manual labeling.
\vspace{-5pt}
\paragraph{Evaluation Fairness and Cluster Pruning.}
Our refinement process may remove clusters that are deemed incoherent during the coherence verification stage. Importantly, clusters are not discarded to optimize evaluation metrics; rather, a cluster is flagged as incoherent only when its generated summary is not semantically supported by representative member texts. This procedure is applied consistently to all clusters produced by each method before evaluation.

To ensure fair comparison, all evaluation metrics are computed on the final cluster structures produced by each method. We explicitly report the number of clusters ($C$) in Table \ref{tab:cluster_coherency} to make structural changes transparent. Notably, baseline refinement methods such as SBERT-based refinement also modify cluster cardinality, indicating that changes in cluster count are not unique to our approach.

Furthermore, improvements are not limited to geometric metrics. As shown in Table \ref{tab:eval}, LLM-based refinement yields substantial gains in human-aligned assignment accuracy, suggesting that improvements reflect better semantic consolidation rather than metric inflation. We also observe that, on the X dataset, the statistical behavior of LLM-refined clusters is comparable to HDBSCAN ($p = 0.48$), and on Bluesky, no method significantly dominates (Table \ref{tab:stat} in App. \ref{app:stat}), indicating that performance gains are not driven by trivial pruning effects.
\subsection{Error Analysis} 
We conduct an error analysis to examine where the best model's (GPT-4o) assigned themes diverge from human judgment. In X, the theme \textit{veganism impacts, challenges, and discussions} overlaps with \textit{advocacy}, \textit{lifestyle}, and \textit{ethics}, causing short personal posts to be misclassified. Food mentions often trigger \textit{dining experiences} labels even for ads or generic content, while positive remarks are sometimes conflated with promotion or activism (e.g., \textit{animal rights}). In Bluesky, abstract themes such as \textit{social and ethical commentary} are frequently misclassified due to implicit moral cues, sarcasm, or vague language. We also observe keyword over-reliance, with mentions of \textit{skincare} or donations misclassified as ethical consumption or advocacy. Error analysis details are provided in App. \ref{app:error}.
\begin{figure}[t]
\includegraphics[width=\columnwidth]{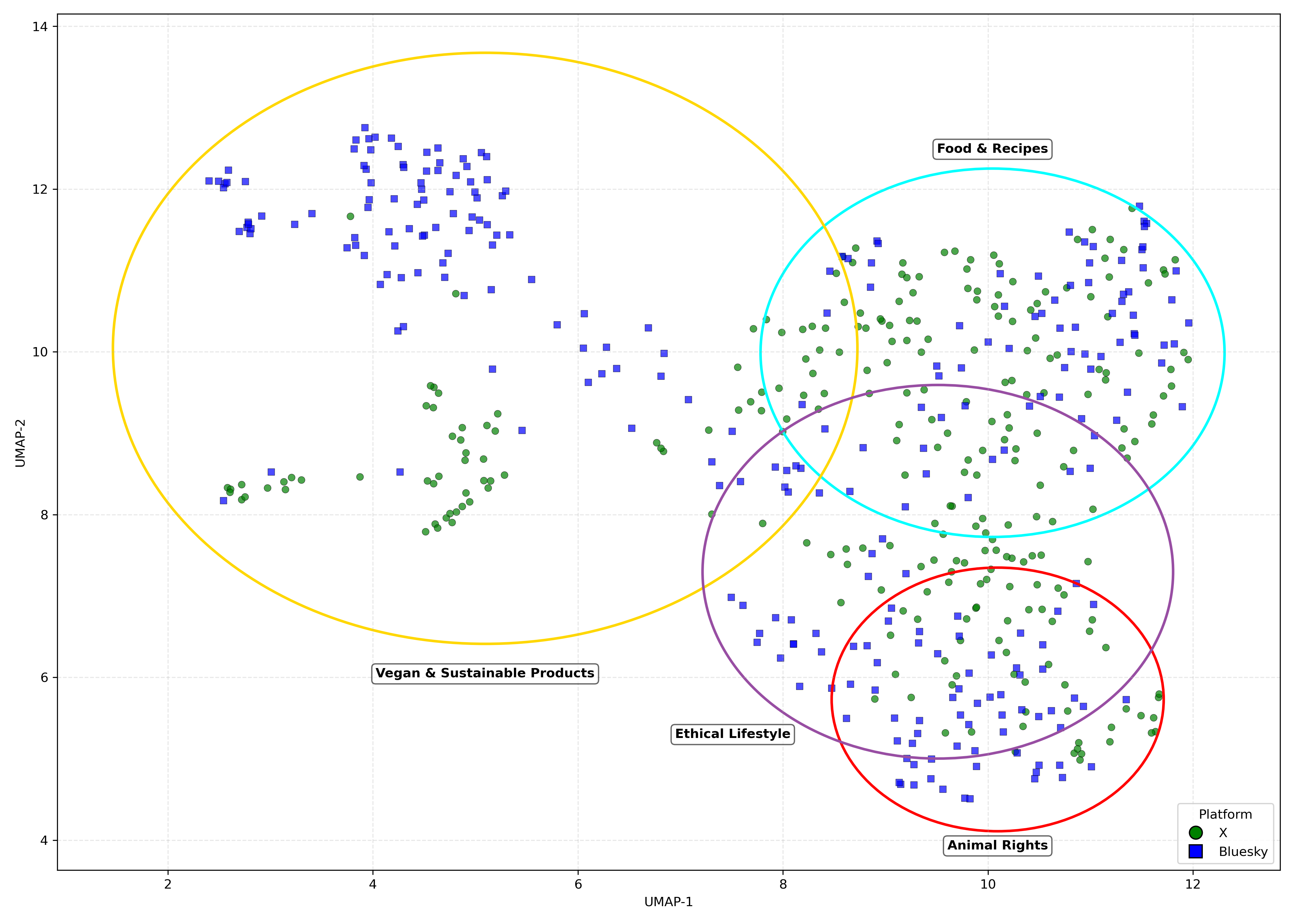}
\caption{{\small UMAP projection of vegan discourse from X (green) and Bluesky (blue) with broader themes.}} 
\vspace{-5pt}
\label{fig:umap_thm}
\vspace{-5pt}
\end{figure}
\begin{table*}
\centering
\small
\begin{tabular}{p{2.6cm}|p{1.2cm}p{11cm}}
\hline
\textbf{Broader Theme} & \textbf{Platform} & \textbf{Example Posts} \\
\hline
\multirow{5}{=}{\raggedright Animal Rights} & \multirow{2}{*}{X} & ...Stop this cruelty! \#AnimalsLivesMatter \#animalcruelty \#vegan \#animalrights PETITION: Justice for Monkey Hanged from Tree ..... \\
\cline{2-3}
                   & \multirow{3}{*}{Bluesky} & Wool industry is infested with violence \& documented cruelty in nearly every shearing shed that investigators entered. One of the shearers is seen punching sheep in the nose and face. Another is seen jabbing a sheep........... \\
\hline
\multirow{3}{=}{\raggedright Food \& Recipes}   & \multirow{2}{*}{X} & ...and I are taco lovers \& all things Mexico so we can't wait to try Veggie Lad's \#vegan recipe...\\
\cline{2-3}
                  & \multirow{2}{*}{Bluesky} & If you want some next-level delicious tofu, try out this vegan orange chicken! \#vegan \#vegansky \#veganfood \#tofu.....\\
\hline
\multirow{4}{=}{\raggedright Vegan \& Sustainable Products}  & \multirow{2}{*}{X} & Why you should be using shampoo bars to have healthier hair and protect the environment \#ecofriendly \#CrueltyFree \#vegan..... \\
\cline{2-3}
                  & \multirow{2}{*}{Bluesky} & Delve into the world of Amazon's organic beauty with eco-friendly, cruelty-free products that redefine skincare \#OrganicBeauty \#SustainableBeauty..... \\

\hline
\multirow{3}{=}{\raggedright Ethical Lifestyle} & X & Celebrate earth day everyday by going vegan...... \\
\cline{2-3}
                  & \multirow{3}{*}{Bluesky} & Veganism doesn’t actually mean zero animal byproduct, it’s about reducing harm to animals so you can in fact be vegan and consume animal byproduct, it just has to be more ethical... \\
\hline
\end{tabular}
\vspace{-5pt}
\caption{Examples of posts by broader theme.}
\vspace{-5pt}
\label{tab:example_content}
\vspace{-5pt}
\end{table*}
\section{Qualitative Analysis}
To determine whether the differences in linguistic tone across the two platforms can be attributed to differing thematic focuses, we map embedding vectors of $1000$ texts from both platforms (same $500$ from previously randomly selected from each platform) onto a 2-dimensional projection using UMAP. \textbf{To enhance clarity, we show broader themes that are closely situated within the same density region} (Fig. \ref{fig:umap_thm}). We observe distinct but overlapping clusters: Bluesky posts dominate the `Vegan \& Sustainable Products' region, while both platforms contribute to `Food \& Recipes', `Ethical Lifestyle', and `Animal Rights'. These latter themes often blend practical advice with advocacy, reflecting how lifestyle choices are framed as moral commitments. The examples of the posts for each broader theme are shown in Table \ref{tab:example_content}. 

Table \ref{tab:unq_thm} in App. \ref{app:thm_dis} highlights example themes that are unique to each platform. On X, discourse centers on informational and aspirational themes (e.g., daily motivation, advocacy), reflecting its broadcast-oriented use. In contrast, Bluesky emphasizes conversational and satirical content, including humor and sociopolitical critique. Theme distributions (Fig.~\ref{fig:thm_dis} in App. \ref{app:thm_dis}) further show that X foregrounds advocacy, while Bluesky leans towards lifestyle and consumer-oriented themes, highlighting platform-specific discourse patterns.

%
\section{Temporal Drift and Robustness}
Our X corpus spans late $2019$–early $2020$, while Bluesky data is from June'25, raising concerns about confounding historical and platform effects. To address this, we identify the densest $28$-day window of X activity ($2020.01.14$ – $2020.02.10$) and down-sample Bluesky to match post count. Cluster quality improves for both platforms under this balanced setup, but the relative pattern sharpens: X shows higher cohesion (silhouette $0.60$ vs.\ $0.08$) and tighter separation (Davies--Bouldin $0.61$ vs.\ $1.14$). A $\chi^2$ test \cite{cochran1952chi2} on the \textit{theme}~$\times$~\textit{platform} table confirms the association is \textbf{non-random} ($\chi^2 = 80.0$, $\mathit{df} = 23$, $p \approx 3 \times 10^{-8}$). Hence, our findings are \textbf{descriptive} signals that persist after equalizing volume and narrowing time window—\textbf{not artifacts of dataset imbalance}. See the App. \ref{app:temporal} for full methodology and visualizations (Fig. \ref{fig:bal}). We emphasize that these are \textbf{descriptive contrasts}—\textbf{not causal estimates of platform design} because external factors (e.g., COVID-19, market shifts) may influence discourse. 

\section{Broader Societal Impact}
This work has meaningful implications for understanding digital public discourse.
Our approach offers a powerful tool for \textit{sociotechnical} analysis. This enables researchers, policymakers, and stakeholders to gain timely insights into understanding trends, evolving public sentiments, and consumer behavior at scale.

Our case study illustrates the importance of analyzing narratives across both centralized (X) and decentralized (Bluesky) platforms. We show that different platforms amplify distinct facets of the same movement—ranging from {\small\texttt{ethical advocacy}} to {\small\texttt{humor}}, {\small\texttt{product marketing}}, and {\small\texttt{community identity}}. These distinctions reflect broader shifts in how digital infrastructure shapes discourse, activism, and ideology. 

Our findings offer actionable insights for stakeholder groups such as vegan advocacy organizations, e.g., PETA\footnote{\url{https://www.peta.org/}}, TVS\footnote{\url{https://www.vegansociety.com/}}, and consumer protection watchdogs, e.g., FTC\footnote{\url{https://www.ftc.gov/}}. For instance, an advocacy group aiming to mobilize support for {\small\texttt{animal rights}} might prioritize Bluesky’s \textit{community-driven environment}, while consumer protection agencies concerned with greenwashing could scrutinize high-volume \textit{product promotions} on X. By highlighting differences in thematic structure and coherence, our framework can inform where and how to engage diverse online audiences.

\section{Conclusion and Discussion}
We introduce a reasoning-based framework that uses LLMs as semantic judges to validate and refine unsupervised text clusters. Rather than inducing structure from scratch, the approach improves semantic coherence and human-aligned labeling while maintaining competitive separation relative to strong embedding-based baselines. Experiments with cross-platform social media data and robustness analysis demonstrate the reliability of the induced structure. 

Our method is particularly beneficial in settings where initial clustering produces semantically noisy or redundant structures, such as short-text or high-variation corpora. When clustering quality is already near-optimal, refinement yields smaller gains, suggesting that the framework is most impactful as a semantic validation layer rather than a replacement for strong clustering methods.

\section{Limitations}
Our experiments focus on English-language social media data and a specific discourse domain (veganism), which may limit direct generalization to other languages or domains. While the framework itself is domain-agnostic, applying it to new settings may require adapting prompts or evaluation criteria. Finally, although human validation demonstrates strong alignment with LLM-based labeling, we \textbf{do not} claim that LLM judgments fully replace expert annotation in high-stakes settings.

In our experiment, as the X data predates Bluesky by $\sim$5 years, our analysis offers descriptive—not causal—contrasts. The 28-day subsample narrows but does not remove this gap—because residual historical confounds (e.g., COVID-19, market shifts) cannot be ruled out without contemporaneous data from both platforms. Future work might collect $2025$ X sample to eliminate this residual confound.

Additionally, as LLMs are trained on extensive human-generated text, they may embed human biases \cite{islam2026gets,blodgett2020language,brown2020language}, which are not addressed in this study.
We only used pre-trained LLMs and did not consider fine-tuning due to the resource constraints. 

We adopt TF–IDF to preserve interpretability and lexical transparency in early clustering stages, though dense embedding models could serve as alternatives.  Future work could replace TF–IDF with dense encoders such as E5 or GTR-T5 to evaluate whether dense initializations further enhance cluster coherence in cross-platform discourse analysis.

We emphasize that our work does not propose a new topic model nor generate topics directly with LLMs; instead, it focuses on post-hoc semantic validation of arbitrary unsupervised clustering methods. Our current evaluation focuses on intra-cluster embedding similarity, Silhouette Score, Davies–Bouldin Index, and human validation. Incorporating document-level coherence metrics \cite{rahimi-etal-2024-contextualized,korenvcic2018document,ramrakhiyani-etal-2017-measuring} adapted for clustering outputs can be explored as future work.



\section{Ethical Considerations}
To the best of our knowledge, we did not violate any ethical code while conducting the research work described in this paper. We report the technical details for the reproducibility of the results. The author's personal views are not represented in any results we report, as it is solely outcomes derived from machine learning or AI models.

The social media data used in this study may contain offensive, biased, toxic, or harmful language. Such content reflects user-generated discourse and does not represent the views of the authors or institutions. All analyses were conducted for research purposes only.

\section{Acknowledgments} 
We would like to thank Lightning AI Studio for providing the computing resources. Also, we are thankful to the anonymous reviewers for their thoughtful suggestions. 


\bibliography{custom}

\appendix

\section{Data Collection}
\label{app:dc}
The full list of keywords can be seen in Table \ref{tab:keywords}.
\begin{table}[h]
\centering
\begin{tabular}{|l|}
\hline
vegandiet, 
veganfood, 
veganlife,
veganlover, \\
veganlifestyle, 
vegancommunity,\\
veganfoodshare,
vegans,
veganfoodlove, \\
veganjourney,
plant-based, 
cruelty-free, \\ 
dairyfree,
crueltyfree,
meat-free, \\
govegan,
veganfoodporn, 
animal rights. \\
     \hline
    \end{tabular}%
\caption{List of the keywords for data collection.} 
\label{tab:keywords}
\end{table}
\section{Experiments}
\subsection{HDBSCAN Hyperparameters}
\label{app:hdbscan}
We grid search for the number of $min\_samples$ on $\{2, 3, 5, 10\}$. The $min\_cluster\_size$ number is selected
based on a grid search whose values are sensitive
to the number of input data points. Suppose $|D|$
denote the number of data points, then the grid
parameters for HDBSCAN used in our method include $\{$5$, $10$, $15$, $0.05$\times |D|, $0.1$\times |D|, $0.2$\times |D|, $0.25$\times|D|\}$. We set
the $n\_neighbors$ parameter in UMAP embedding
to $min\_cluster\_size$. 
In our framework, for X, the best parameters are $min\_cluster\_size: 15$, $min\_samples: 3$ to obtain the DBCV score of $0.35$. For Bluesky, the best parameters are $min\_cluster\_size: 10$, $min\_samples: 2$ to obtain the DBCV score of $0.53$.
\subsection{Cluster Threshold Selection}
\label{app:threshold_selection}
We grid search the merge similarity threshold $\tau \in \{0.75, 0.80, 0.85, 0.90\}$ and evaluate each setting using Silhouette Score ($S$), Davies–Bouldin Index ($DB_i$), and the resulting number of clusters ($C$). $S$: higher means points fit their own cluster better than others. $DB_i$: lower means clusters are compact and well separated. Results are shown in Table \ref{tab:thrs_grid}. While quality metrics improve with higher $\tau$, $0.90$ yields an explosion in small, redundant clusters that hurt interpretability and summarization. We therefore select $0.85$ as the best trade-off—strong scores with controlled cluster count.
\begin{table}
\centering
\begin{tabular}{p{0.1\textwidth} | p{0.25\textwidth}}
\hline
Dataset & $\tau$ $(C, S, DB_i)$ \\
\hline
\multirow{2}{*}{X} & 0.75 (126, 0.378, 1.037), 
0.80 (174, 0.519, 0.838), \\
& \textbf{0.85 (232, 0.674, 0.635)}, 
0.90 (292, 0.825, 0.493)  \\
\hline
\multirow{2}{*}{Bluesky} & 0.75 (22, 0.589, 0.852), 
0.80 (32, 0.859, 0.498), \\
& \textbf{0.85 (36, 0.979, 0.227)}, 
0.90 (37, 1.0, $6.2e-08$)\\
\hline
\end{tabular}
\caption{Threshold selection using grid search.} 
\label{tab:thrs_grid}
\end{table}
\subsection{Prompt Design}
\label{app:prompt}
\begin{figure}[h]
\includegraphics[width=\columnwidth]{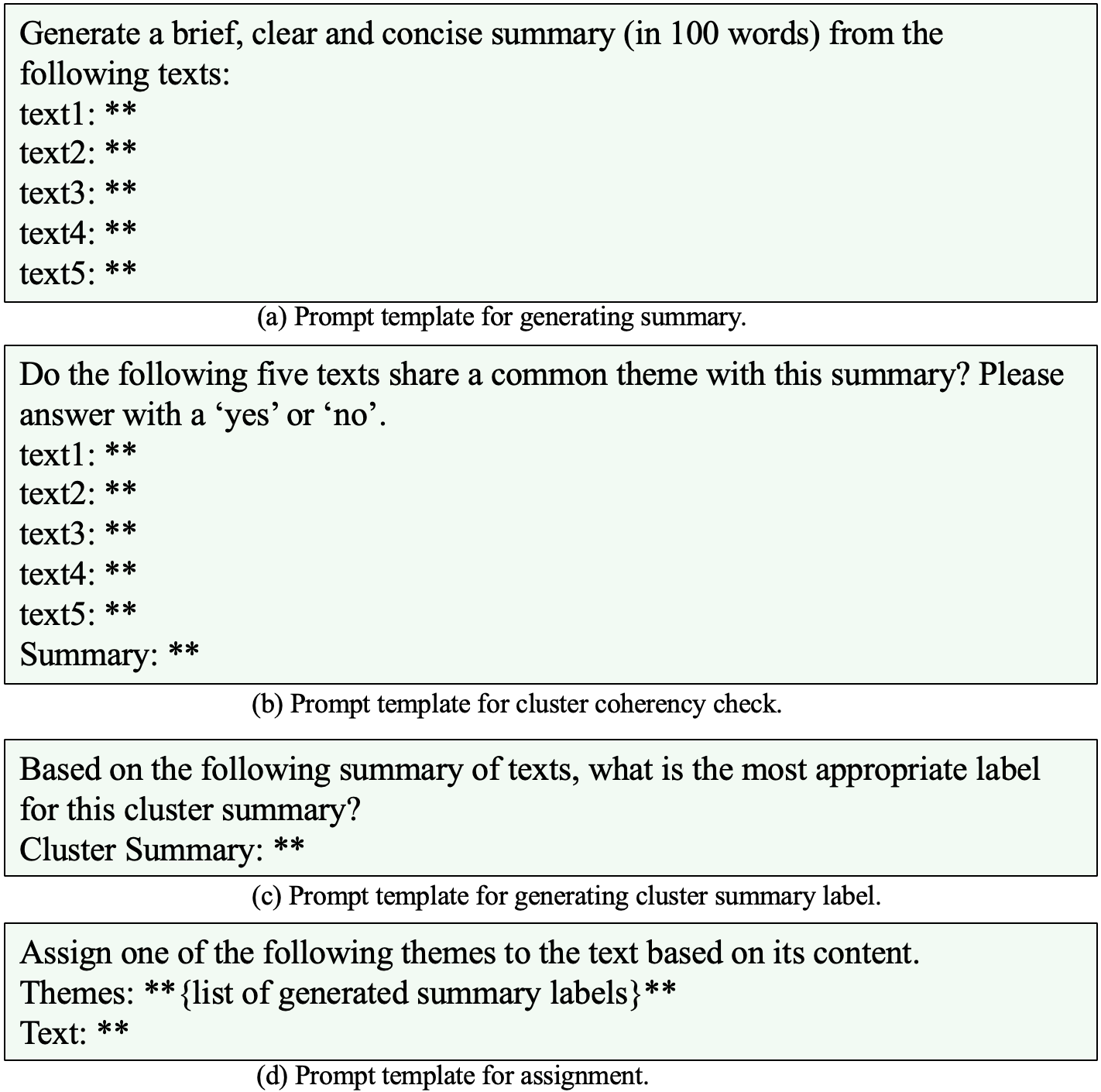}
\caption{Prompt templates (shown as zero-shot).} 
\label{prompt_template}
\end{figure}

\begin{figure*}
\includegraphics[width=\textwidth]{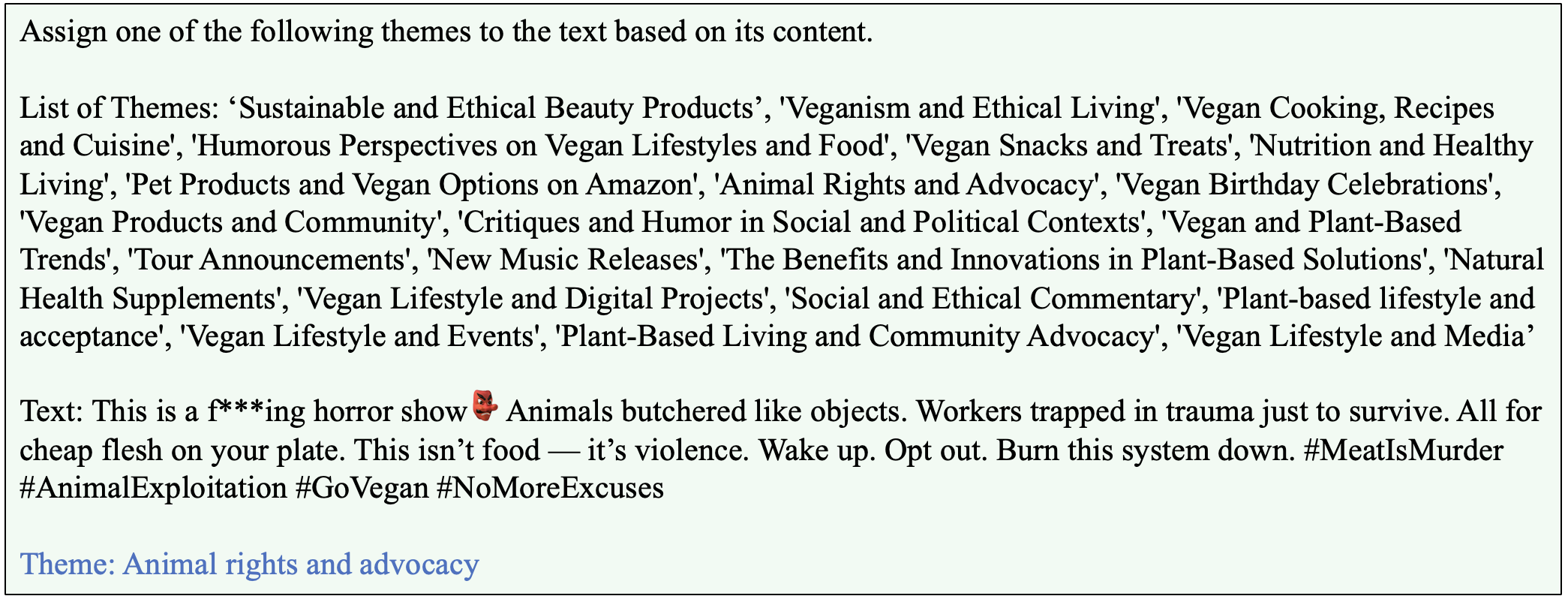}
\caption{Prompt example of mapping text$\rightarrow$theme (Bluesky dataset). The \textit{black} colored segment is the \textit{input} prompt and the \textit{blue} colored segment is the generated \textit{output} by LLMs.} 
\label{prompt_ex}
\end{figure*}

Prompt templates used in our work are illustrated in Fig. \ref{prompt_template}. To summarize the top-5 texts, the prompt template is shown in Fig. \ref{prompt_template}(a). Fig. \ref{prompt_template}(b) provides a prompt template for checking cluster coherency, and Fig. \ref{prompt_template}(c) represents a prompt template for generating a cluster summary label. Prompt template for assigning a label (from the list of generated summary labels) to individual text is shown in Fig. \ref{prompt_template} (d). Fig. \ref{prompt_ex} shows the prompt example of assigning text to the label.
\begin{table*}
\centering
\begin{tabular}{p{0.07\textwidth} | p{0.85\textwidth}}
\hline
\textbf{Dataset} & \textbf{Labels} \\
\hline
\multirow{6}{*}{X} & Daily Motivation and Inspiration, Gluten-Free and Vegan Publication Updates, 
Veganism Advocacy and Lifestyle Promotion, Vegan and Vegetarian Recipes and Cookbook, 
Veganism and Plant-Based Ethical Lifestyle, Veganism Advocacy and Animal Rights, Vegan Food and Lifestyle Celebrations, 
Vegan Desserts Recipe, Handmade Vegan Soaps Promotion, 
Veganism Impacts, Challenges, and Discussions, 
Vegan and Vegetarian Dining Experiences, 
Vegan and Gluten-Free Food Promotions, 
Promotion of Low-Sugar Vegan Tea Products, 
Promotion of Vegan Haircare Products. \\
\hline
\multirow{7}{*}{Bluesky} & Sustainable and Ethical Beauty Products, Veganism and Ethical Living, 
Vegan Cooking, Recipes and Cuisine, Humorous Perspectives on Vegan Lifestyles and Food, 
Vegan Snacks and Treats, Nutrition and Healthy Living, 
Pet Products and Vegan Options on Amazon, 
Animal Rights and Advocacy, Vegan Birthday Celebrations, 
Vegan Products and Community, 
Critiques and Humor in Social and Political Contexts, 
Vegan and Plant-Based Trends, Tour Announcements, New Music Releases, 
The Benefits and Innovations in Plant-Based Solutions, 
Natural Health Supplements, Vegan Lifestyle and Digital Projects, 
Social and Ethical Commentary, Plant-based lifestyle and acceptance, 
Vegan Lifestyle and Events, Plant-Based Living and Community Advocacy, 
Vegan Lifestyle and Media. \\
\hline
\end{tabular}
\caption{Generated labels by LLMs from X and Bluesky data.}
\vspace{-5pt}
\label{tab:label}
\end{table*}
\begin{figure*}
\includegraphics[width=\textwidth]{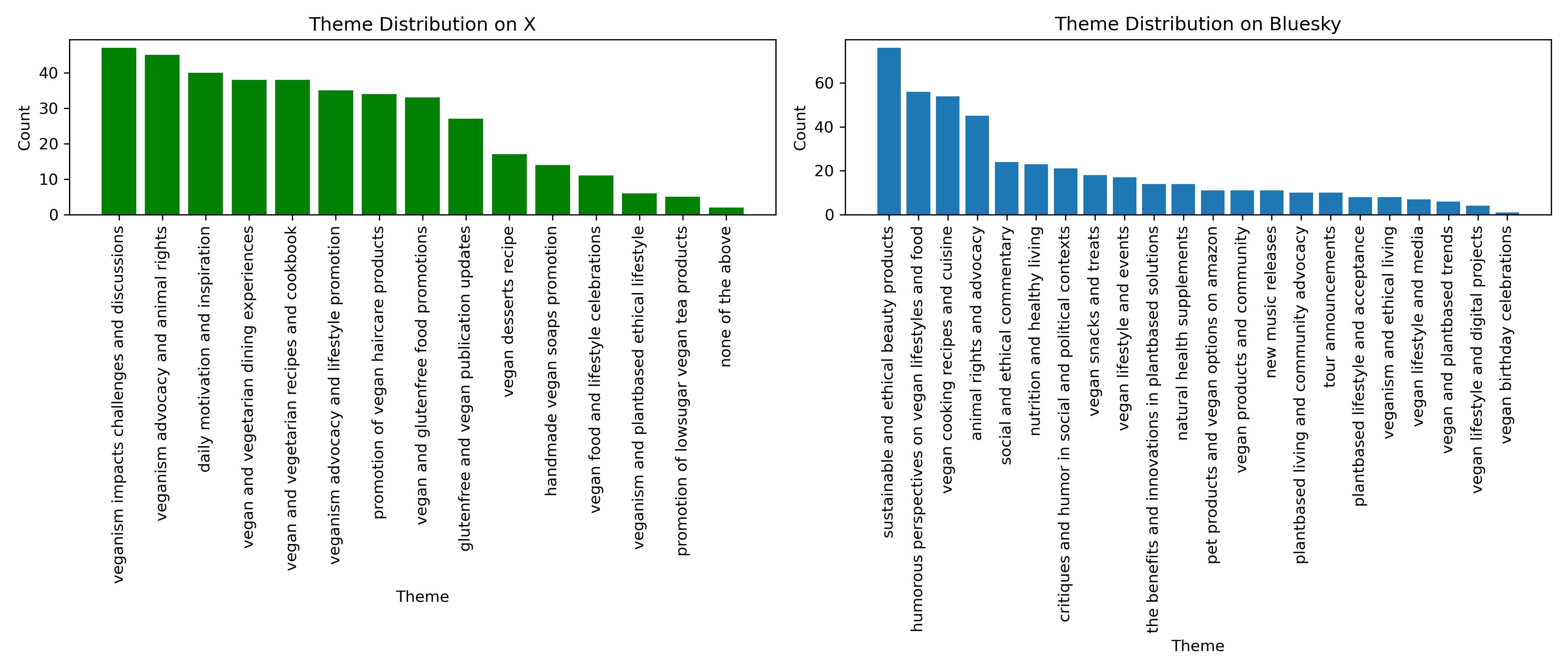}
\caption{Distribution of assigned labels/themes using GPT-4o on X (green) and Bluesky (blue) datasets.} 
\label{fig:thm_dis}
\end{figure*}
\begin{table*}
\centering
\renewcommand{\arraystretch}{1.3}
\begin{tabular}{
>{\centering\arraybackslash}m{0.07\textwidth} |
>{\raggedright\arraybackslash}m{0.35\textwidth} |
>{\centering\arraybackslash}m{0.15\textwidth} |
>{\centering\arraybackslash}m{0.15\textwidth} |
>{\centering\arraybackslash}m{0.14\textwidth}}
\hline
\textbf{Dataset} & \textbf{Test / Comparison} & \textbf{Statistic} & \textbf{p-value} & \textbf{Significance} \\
\hline

\multirow{6}{*}{X} 
 & Kruskal--Wallis H-statistic & 16.187 & 0.0003 & $p<0.001$ \\
\cline{2-5}
 & Mann--Whitney U: HDBSCAN vs LLM & 42706.0 & 0.4808 & n.s. \\
 & Mann--Whitney U: HDBSCAN vs SBERT & 52922.5 & 0.000165 & $p<0.001$ \\
 & Mann--Whitney U: SBERT vs LLM & 24018.0 & 0.001829 & $p<0.01$ \\
\hline
\hline
\multirow{6}{*}{Bluesky} 
 & Kruskal-Wallis H-statistic & 0.2809 & 0.8690 & n.s. \\
\cline{2-5}
 & Mann--Whitney U: HDBSCAN vs LLM & 680.0 & 0.8816 & n.s. \\
 & Mann--Whitney U: HDBSCAN vs SBERT & 674.5 & 0.6044 & n.s. \\
 & Mann--Whitney U: SBERT vs LLM & 582.0 & 0.7289 & n.s. \\
\hline
\end{tabular}
\caption{Statistical test results for cluster metric comparisons. n.s. = statistically not significant (p-value $> 0.05$).}
\label{tab:stat}
\end{table*}
\subsection{Generated Labels}
\label{app:gen_label}
Table \ref{tab:label} shows the generated labels by our framework.
\subsection{Theme Distribution Analysis}
\label{app:thm_dis}
Fig. \ref{fig:thm_dis} presents a comparative analysis of the distribution of assigned themes (generated by GPT-4o) across two social media platforms—X (formerly Twitter) and Bluesky—as part of our study on vegan-related discourse.

\paragraph{Platform X (Left Bar Chart).}
The discourse on X is concentrated around a few dominant themes: `Veganism impacts, challenges and discussions', `Veganism advocacy and animal rights', `Daily motivation and inspiration'.
These themes reflect a strong focus on ideological, motivational, and advocacy-related content, aligning with X’s identity as a platform for public debate and activism. Although there is some thematic diversity, long-tail topics such as \textit{Vegan birthday celebrations} or \textit{None of the above} appear infrequently.

\paragraph{Bluesky (Right Bar Chart).}
Bluesky shows a more top-heavy distribution, with a focus on lifestyle and product-related themes: `Sustainable and ethical beauty products', `Humorous perspectives on vegan lifestyles and food', `Vegan cooking recipes and cuisine'. This indicates that Bluesky users engage more with practical, day-to-day content rather than advocacy or ideological discussions. The themes also display greater granularity, including niche interests such as \textit{Tour announcements}, \textit{Pet products}, and \textit{Plant-based lifestyle and acceptance}.

These differences suggest distinct discursive cultures:
\begin{itemize}
    \item \textbf{X} favors advocacy-driven and community-mobilizing narratives.
    \item \textbf{Bluesky} promotes a lifestyle-oriented discourse, with greater emphasis on ethical consumerism and humor.
\end{itemize}
Such findings support the hypothesis that platform audience composition significantly influences the thematic framing of social media discourse.

Table \ref{tab:unq_thm} shows the themes that are unique to each platform. On X, posts often center on `Daily Motivation and Inspiration' and `Gluten‑Free Vegan Publication Updates', reflecting its role as a hub for {\small\textit{informational and aspirational content}}. In contrast, Bluesky features `Humorous Perspectives on Vegan Lifestyles' and `Critiques in Social and Political Contexts', showing a more {\small\textit{conversational and satirical}} tone.
From theme distributions (Fig. \ref{fig:thm_dis}), we notice that X emphasizes advocacy and motivational discourse (e.g., `Veganism advocacy and animal rights'), while Bluesky leans toward lifestyle and consumer themes (e.g., `Sustainable beauty products', `Vegan recipes'). These distinctions underscore how each platform fosters different facets of vegan discourse.
\begin{table*}
\centering
\small
\begin{tabular}{p{1.2cm}p{5cm}p{8.5 cm}}
\hline
\textbf{Platform} & \textbf{Unique Theme} & \textbf{Example Posts} \\
\hline
\multirow{3}{*}{X} & \multirow{3}{=}{\raggedright Daily Motivation and Inspiration} & 5 BEST EXERCISES FOR PEOPLE w/ BACK PAIN \#fitness \#nutrition \#vegan \#mindfulness \#motivation \#inspiration \#SaturdayMotivation.....  \\
\hline
\multirow{2}{*}{X} & Gluten-Free and Vegan Publication Updates  & @TheHecticVegan: The Hectic Vegan Magazine Issue 5 \#glutenfree \#vegan \\
\hline
\hline
\multirow{2}{*}{Bluesky} & Humorous Perspectives on Vegan Lifestyles and Food & Vegan chefs consider it a huge insult when you cut yourself \& drip your own blood on to their burgers to enhance the flavor. \\
\hline
\multirow{3}{*}{Bluesky} & \multirow{3}{=}{\raggedright Critiques and Humor in Social and Political Contexts} & The funny thing about vegans is that they think they're morally superior for not eating meat because of deforestation when their entire diet is based on exploiting farmers get f****d (moral) vegans. \\
\hline
\end{tabular}
\caption{Examples of platform-exclusive themes and representative posts.}
\vspace{-5pt}
\label{tab:unq_thm}
\end{table*}
\begin{table*}
    \centering
    \begin{tabular}{p{1.3cm}|p{5cm}|p{8.5cm}}
        \hline
        \textbf{Platform} & \textbf{Theme}   & \textbf{Example Posts}\\ 
        \hline
        \multirow{6}{*}{X}  & \multirow{6}{=}{\raggedright veganism impacts, challenges, and discussions}  & .... y’all all the f**king vegan stables are gone. no rice, no beans, no bread, no chips, no oat milk, no kombucha, no peanut.  \\
        \cline{3-3}
        & & I keep striking out in the meat sections of my local grocery stores. This pandemic just might force me into veganism. https://t.co/EwsFYindAn \\
        \hline
        \multirow{4}{*}{X} & \multirow{4}{=}{\raggedright vegan and vegetarian dining experiences}  & ... I’d prob cry if anyone ever cooked vegan for me. \\
        \cline{3-3}
        & &  We look forward to hosting our vegetarian cooking class tonight! See you at 18h00! \#FreshEarth \#CookingClasses http://t.co/WpSz5Uqfki \\
        \hline
        \multirow{6}{*}{X} & \multirow{6}{=}{\raggedright veganism advocacy and lifestyle promotion}  & People who act like they can’t enjoy vegetarian meals because they eat meat are so annoying.  \\
        \cline{3-3}
        & & @Sydney843 @QUBFoodProf @DiscoStew66 Organic vegetables almost always use animal byproducts in fertilisers and therefore are not vegan. Vegans should only ever eat non-organic unless from explicitly from a vegan farm. \\
        \hline
        \hline
        \multirow{6}{*}{Bluesky}  & \multirow{6}{=}{\raggedright social and ethical commentary}  & If only humans were rational beings ... \#plantbased theconversation.com/why-theres-a...  \\
        \cline{3-3}
        & & vegans for sure but for second place it def depends i find mostly other people mention the height unless it’s someone under the age of like 23 and it for the bi/mulit-racial depends on the mix and which parent is which lmao \\
        \hline
        \multirow{10}{*}{Bluesky}  & \multirow{10}{=}{\raggedright sustainable and ethical beauty products}  & HybridGel Fill  \#EnailCouture  \#NailzByDragon \#DragonzClaw \#VeganNailTech \#NailTech \#VeganNails \#208VeganNailTech \#BoiseNailTech \#BoiseNails \#BoiseIdaho \#UniqueNails \#HybridGelNails \#AcrylicNails \#Manicure \#Pedicure \#GelManicure \#GelPedicure \#HybridGel \#Polygel \#HappyGel \#AcrylGel \\
        \cline{3-3}
        & & Seven of Limes is back! This \#StarTrekDiscovery -inspired soap was a big hit last year. Watch these citrusy swirls all come together here: \url{youtu.be/hcJ93curg_0?} \#soapmaking \#vegansoap \#howitsmade \#sevenoflimes \\
        \hline
        \multirow{5}{*}{Bluesky}  & \multirow{5}{=}{\raggedright animal rights and advocacy}  & @afivegantenna.bsky.social Pls my friend donate if you can and Share the post with the donation link. bsky.app/profile/9ahm \\
        \cline{3-3}
        & & @liberalvegan.bsky.social bsky.app/profile/abed... Please, my friend, donate if you can and write a quote. \\
        \hline
    \end{tabular}
    \caption{Example posts from X and Bluesky where GPT-4o's assigned themes \textbf{do not} align with human judgment.}
    \label{tab:error}
\end{table*}
\subsection{Experimental Cost}
\label{app:cost}
Overall, in our experiment, for X data, GPT-4o costs $\approx \$22$, and for Bluesky data, GPT-4o costs  $\approx \$14$. Latency (per $1k$ posts) $\approx$ $6-10$ minutes.

\section{Statistical Tests}
\label{app:stat}
Table \ref{tab:stat} summarizes the statistical significance analysis of the clustering results for both the X and Bluesky vegan datasets. We first apply a non‑parametric Kruskal–Wallis test to determine whether there are overall differences in the clustering quality metrics (Silhouette Score and Davies–Bouldin Score). Then we do pairwise comparisons (independent samples) using the Mann–Whitney U test. 
\section{Error Analysis Details}
\label{app:error}
For X, the top-$3$ most frequent themes associated with errors are as follows: `veganism impacts, challenges, and discussions' ($12$ times), `vegan and vegetarian dining experiences' ($11$ times), `veganism advocacy and lifestyle promotion' ($11$ times). For Bluesky, the top-$3$ most frequent themes associated with errors are as follows: `social and ethical commentary' ($8$ times), `sustainable and ethical beauty products' ($4$ times), `animal rights and advocacy' ($4$ times). 
In Table \ref{tab:error}, we show a few cases where the themes assigned by GPT-4o do not align with human evaluations. 
\subsection{Error Analysis: X Data}
Possible reasons for misclassifications of `veganism impacts, challenges, and discussions':
This theme might overlap with others like \textit{advocacy}, \textit{lifestyle}, and \textit{ethics}. The model may be assigning this category when the content is not analytical or discussion-based, e.g., a short personal story.

Possible reasons for misclassifications of `vegan and vegetarian dining experiences':
Confusion between promotion and personal experience.
Overgeneralization: GPT might label any mention of food or restaurants under this, even if the content is not about personal experiences (e.g., advertisements or random food mentions). 
False positive on food references: Posts like {\small\texttt{I’d prob cry if anyone ever cooked vegan for me}} or {\small\texttt{hosting vegan cooking class}} should not qualify as a \textit{dining experience}.

Possible reasons for misclassifications of `veganism advocacy and lifestyle promotion':
GPT-4o may assign this theme to any positive vegan content, even if it's not clearly promotional or advocacy-related. Posts like {\small\texttt{People who act like they can’t enjoy vegetarian meals because they eat meat are so annoying}}. On the other hand, the model might confuse with more cause-driven themes like animal rights or activism, such as posts like {\small\texttt{@Sydney843 @QUBFoodProf @DiscoStew66 Organic vegetables almost always use animal byproducts in fertilisers and therefore are not vegan. Vegans should only ever eat non-organic unless from explicitly from a vegan farm}}. 

Additionally, for very few instances, LLMs provide a `None of the above' response when a user posts about an unrelated topic but includes the hashtag \#vegan. 
\subsection{Error Analysis: Bluesky Data}
The theme `social and ethical commentary' is prone to misclassifications due to its abstract and context-dependent nature. Posts under this theme often involve implicit moral reasoning, sarcasm, or indirect critique, which GPT-4o may struggle to interpret without deeper discourse understanding. For instance, vague or philosophical remarks like {\small\texttt{If only humans were rational beings}} can be easily mistaken for general lifestyle reflections or advocacy, especially when accompanied by hashtags such as \#plantbased. Moreover, the absence of explicit topical anchors (e.g., product names, events, or actions) makes it difficult for the model to confidently associate the content with ethical or societal discourse. 

We notice that posts mentioning skincare, gel, or beauty are classified as `sustainable and ethical beauty products', showing the model’s keyword over‑reliance—ads and self‑care tips with no ethical or sustainability angle. 

Another four errors involved with `animal rights and advocacy' where the model treats any emotional plea or donation link (e.g., {\small\texttt{…please, my friend, donate if you can}}) as full‑blown activism, conflating charity appeals with broader advocacy. 

Moreover, posts dominated by hashtags with minimal textual context also led to misclassifications; context-light, hashtag-heavy content encouraged overreliance on surface cues. 

\section{Temporal Drift \& Robustness Analysis}
\label{app:temporal}
We quantify temporal drift by plotting monthly volume and correlating LLM-assigned themes with posting month (Table \ref{tab:x_cor}). Results confirm that several themes (e.g., `vegan hair-care promotions', `veganism advocacy and animal rights') peaked in late 2019, whereas Bluesky’s June'25 content focuses on `plant-based ethics'. 
\begin{table*}
\centering
\begin{tabular}{l|c}
\hline
\textbf{Theme} & \textbf{pearson\_corr\_with\_month}  \\ \hline
Promotion of vegan haircare products              & 0.771 \\
Veganism advocacy and animal rights               & 0.560 \\
Veganism advocacy and lifestyle promotion         & 0.495 \\
Vegan and glutenfree food promotions              & 0.459 \\
Veganism impacts challenges and discussions       & 0.440 \\
Vegan and vegetarian dining experiences           & 0.352 \\
Handmade vegan soaps promotion                    & 0.350 \\
Vegan and vegetarian recipes and cookbook         & 0.277 \\
Veganism and plantbased ethical lifestyle         & 0.250 \\
Glutenfree and vegan publication updates          & 0.232 \\
Vegan food and lifestyle celebrations             & 0.226 \\
Vegan desserts recipe                             & 0.185 \\
Daily motivation and inspiration                  & 0.136 \\
Promotion of lowsugar vegan tea products          & 0.070 \\
\hline
\end{tabular}
\caption{Theme–time correlation on X.}
\label{tab:x_cor}
\end{table*}

Because our X corpus covers Oct'19–Feb'20 while the Bluesky crawl is restricted to June'25, direct comparison risks conflating platform and historical effects. To gauge the impact of this mismatch, we extract the densest $28$-day window of X activity ($2020.01.14$ – $2020.02.10$) by applying a rolling $28$-day sum over daily post counts and selecting the peak. We then down-sample Bluesky to the same number of posts. Next, we recompute the clustering quality for each platform.

Table \ref{tab:full_bal_qual} shows that the relative pattern persists—and even sharpens. X remains far more cohesive than Bluesky: silhouette score increases from $0.40 \rightarrow 0.60$ for X but remains low for Bluesky ($0.15 \rightarrow 0.08$). Separation improves once volume is equalized. Davies-Bouldin scores drop substantially for both platforms ($\approx 3 \rightarrow 1$), yet X still exhibits tighter clustering ($0.61 < 1.14$). Differences are not driven by sheer data volume or longer time span on X; instead, they reflect how discourse is organized.
\begin{table*}
\centering
\begin{tabular}{llrrrr}
\toprule
\textbf{Corpus} & \textbf{Platform} & \textbf{\#Posts} & \textbf{\#Clusters} & \textbf{Silhouette ($\uparrow$)} & \textbf{Davies--Bouldin ($\downarrow$)} \\
\midrule
\multirow{2}{*}{Full} & X        & 939  & 14 & 0.40 & 3.12 \\
& Bluesky  & 2479 & 21 & 0.15 & 3.06 \\
\midrule
\multirow{2}{*}{Balanced 28-day snapshot} 
& X  & 40   & 7  & 0.60 & 0.61 \\
& Bluesky  & 40  & 17 & 0.08 & 1.14 \\
\bottomrule
\end{tabular}
\caption{Cluster quality metrics by platform and corpus.}
\label{tab:full_bal_qual}
\end{table*}

\begin{figure*}
\includegraphics[width=\textwidth]{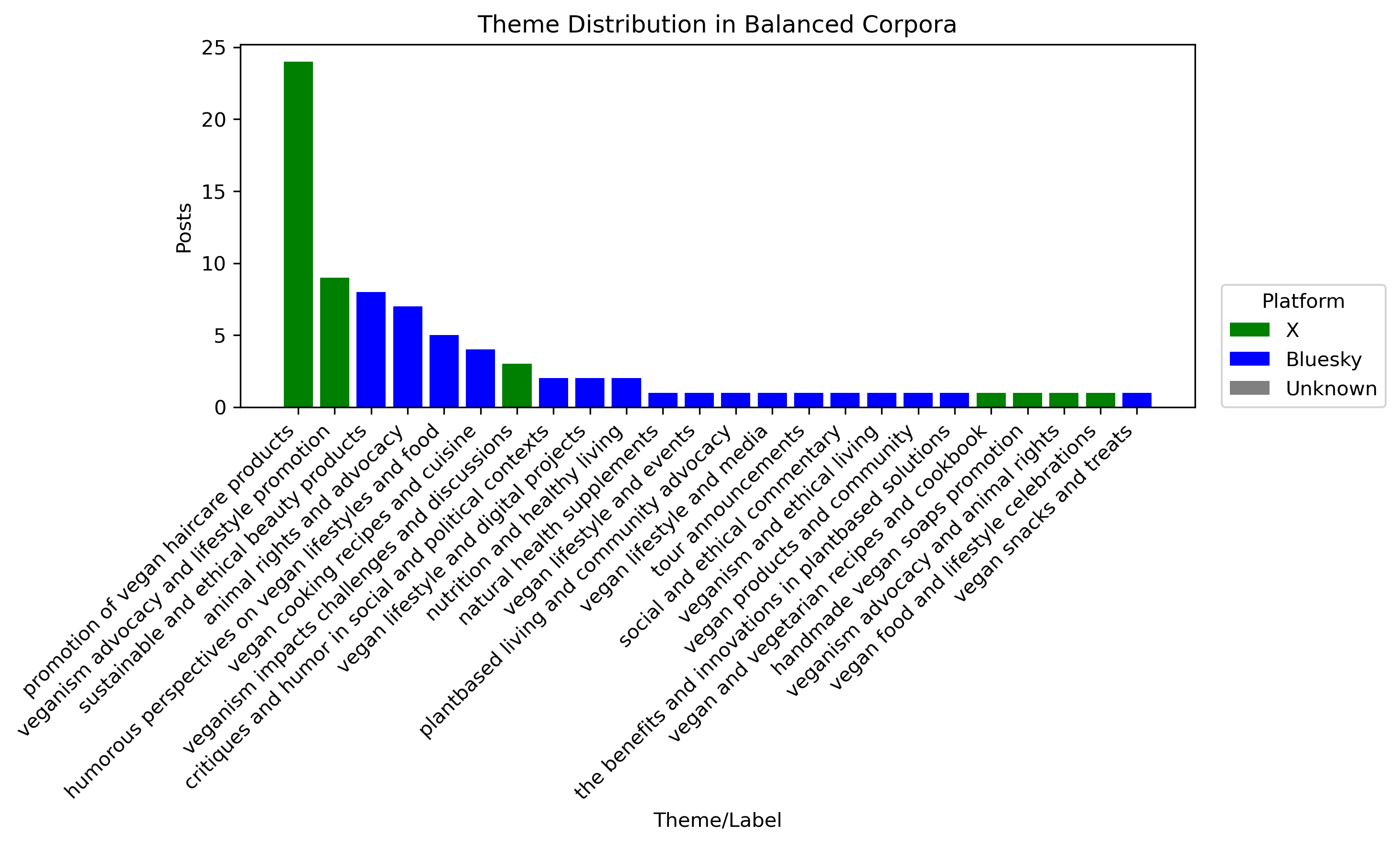}
\caption{Theme distribution in balanced corpora.} 
\label{fig:bal}
\end{figure*}

We plot the distribution of themes (Fig. \ref{fig:bal}) in the balanced snapshot. We visualize the topical mix in the balanced snapshot by counting posts per GPT-4o assigned theme and coloring bars by platform (X vs. Bluesky); totals are equal across platforms by construction. Fig. \ref{fig:bal} shows a strongly heavy-tailed (Zipf-like) distribution: a single promotional theme—`promotion of vegan hair-care products'—dominates the X slice, while Bluesky contributions are diffuse across several themes such as `sustainable \& ethical beauty products', `animal rights \& advocacy', and `humorous perspectives on vegan lifestyles'. In other words, even after equalizing volume and narrowing the time window, the platforms exhibit different topical mixes: X concentrates attention in one large, commerce-oriented theme; Bluesky spreads attention across multiple smaller, lifestyle/advocacy themes. This pattern is consistent with our qualitative reading and aligns with the per-platform cohesion scores: X’s large promotional cluster yields higher within-theme lexical cohesion, whereas Bluesky’s broader mix produces many small clusters. We treat these as \textbf{descriptive contrasts}, \textbf{not causal claims about platform design}.

Then we perform a chi-square test \cite{cochran1952chi2} on the theme$\times$platform contingency table. A $\chi^2$ test of independence on the \textit{theme}~$\times$~\textit{platform} contingency table confirms a non-random association between discourse themes and platform in the balanced snapshot ($\chi^2 = 80.0$, $\mathit{df} = 23$, $p \approx 3.18 \times 10^{-8}$).

\end{document}